\pdfoutput=1

\documentclass[11pt]{article}

\usepackage[final]{acl}

\usepackage{times}
\usepackage{latexsym}
\usepackage{multirow}
\usepackage{graphicx}
\usepackage{enumitem}
\usepackage{adjustbox}
\usepackage{booktabs}
\usepackage{amsmath}
\usepackage{linguex}

\usepackage{arydshln}
\usepackage{hyperref}
\usepackage[T1]{fontenc}

\usepackage[utf8]{inputenc}

\usepackage{microtype}

\usepackage{inconsolata}

\usepackage{graphicx}
\usepackage{xcolor}

\definecolor{green}{HTML}{99CA11}

\title{Beyond Memorization: Assessing Semantic Generalization in Large Language Models Using Phrasal Constructions}

\author{%
\textbf{Wesley Scivetti$^1$\thanks{Equal Contribution},
Melissa Torgbi$^2$\footnotemark[1],
Austin Blodgett$^3$,
Mollie Shichman$^4$,
Taylor Pellegrin$^3$,} \\
\textbf{Claire Bonial$^3$, Harish Tayyar Madabushi$^2$}
\\
$^1$Georgetown University, 
$^2$University of Bath, \\
$^3$DEVCOM U.S. Army Research Laboratory, 
$^4$University of Maryland, College Park \\
\texttt{wss37@georgetown.edu}, \texttt{mat66@bath.ac.uk}
}

\begin{document}
\maketitle
\begin{abstract}
The web-scale of pretraining data has created an important evaluation challenge: to disentangle linguistic competence on cases well-represented in pretraining data from generalization to out-of-domain language, specifically the dynamic, real-world instances less common in pretraining data. To this end, we construct a diagnostic evaluation to systematically assess natural language \emph{understanding} in LLMs by leveraging Construction Grammar (CxG). CxG provides a psycholinguistically grounded framework for testing generalization, as it explicitly links syntactic forms to abstract, non-lexical meanings. Our novel inference evaluation dataset consists of English phrasal constructions, for which speakers are known to be able to abstract over commonplace instantiations in order to understand and produce creative instantiations. Our evaluation dataset uses CxG to evaluate two central questions:  first, if models can `understand' the semantics of sentences for instances that are likely to appear in pretraining data less often, but are intuitive and easy for people to understand. Second, if LLMs can deploy the appropriate constructional semantics given constructions that are syntactically identical but with divergent meanings. 
Our results demonstrate that state-of-the-art models, including GPT-o1, exhibit a performance drop of over 40\% on our second task, revealing a failure to generalize over syntactically identical forms to arrive at distinct constructional meanings in the way humans do. We make our novel dataset and associated experimental data, including prompts and model responses, publicly available.\footnote{\href{https://github.com/melissatorgbi/beyond-memorization}{https://github.com/melissatorgbi/beyond-memorization}}
\end{abstract}

\section{Introduction}

\begin{table}[]
\begin{tabular}{llll}
\toprule
\textbf{Model}       & \multicolumn{2}{l}{\begin{tabular}[c]{@{}l@{}}\textbf{Constructional}\\ \textbf{Semantics}\end{tabular}} & \begin{tabular}[c]{@{}l@{}}\textbf{Construction}\\ \textbf{Distinction}\end{tabular} \\

\midrule
GPT-4o      & 0.88                                    &                                     & 0.58                                                        \\
GPT-o1      & 0.90                                    &                                     & 0.46                                                        \\
Llama 3 70B & 0.74                                    &                                    & 0.52   \\
\hdashline
Human  & 0.90 &  & 0.83\\
\end{tabular}
\caption{We demonstrate a drop in performance, even in the latest models, as we move from evaluating functional understanding of constructional semantics to understanding syntactically identical but semantically distinct constructions. We report accuracy on NLI tasks leveraging distinct constructional premises.} 
\label{tab:overallresults}
\end{table}

Understanding the extent to which Large Language Models (LLMs) generalize from relatively frequent phenomena well-represented in pretraining data to creative, novel usages of language has important implications for LLM development. Identifying the precise nature and limits of LLM generalization can inform decisions about architectures and training regimes~\cite{10.5555/3618408.3619217,zhang2023doesincontextlearninglearn}. This becomes especially relevant as models move toward `reasoning'-based systems and the inevitable widespread deployment of AI agents~\cite{deepseekai2025deepseekr1incentivizingreasoningcapability}. Identifying failure patterns will enable targeted improvements at different stages of development. However, testing LLMs' ability to generalize is particularly challenging because they are trained on vast web-scale data~\cite{lu2024emergentabilitieslargelanguage}. Even if pretraining datasets were fully accessible, ensuring that a test example is truly independent would remain nontrivial, as a model may not have encountered that instance but could have been exposed to related cases that provide indirect information~\cite{madabushi2025stochasticparrotingagillms}. 
\begin{table*}[t]
\centering
\resizebox{\textwidth}{!}{%
\begin{tabular}{lll}
                                                 & \textbf{Exp. 1 Dataset: CxNLI}                     & \textbf{Exp. 2 Dataset: CxNLI-Distinction}                            \\ 
\textbf{CxNLI Premise}      & I brushed my hair smooth.                  & A famous emperor buried scholars alive.           \\
\textbf{Entrenched Variant} & I made my hair smooth (by brushing).       & \textit{NA: no entrenched variants for Exp2 Cxns} \\
\textbf{Cxn Form}           & NP1 V NP2 ADJ                              & NP1 V NP2 ADJ                                     \\
\textbf{Cxn Meaning}        & NP1's action of V causes NP2 to become ADJ & NP2 is ADJ during NP1's action of V               \\
\textbf{CxNLI Hypothesis}   & My brushing caused my hair to be smooth.   & Burying the scholars caused them to be alive.     \\
\textbf{CxNLI Relation}     & Entailment                                 & Contradiction                                    
\end{tabular}%
}
\caption{Example items illustrating experiments. Exp. 1 uses Natural Language Inference (NLI) to test abstraction of the semantics of frequent, entrenched Cxns (usually realized with the verb of the Entrenched Variant) when realized as a creative, infrequent instantiation (CxNLI Premise). Exp. 2 uses NLI to test if models can distinguish and apply the appropriate, distinct semantic interpretation of Cxns that are syntactically identical to the entrenched Cxns of Exp. 1 (with grammatical phrase types of Noun Phrase (NP), Verb (V), a second NP, and an Adjective (ADJ)). }
\label{tab:cxnli_exp1_exp2}
\end{table*}

Therefore, this work introduces a novel evaluation dataset grounded in the theory of Construction Grammar (CxG) \citep{Goldberg_1995,Croft_2001} (see Appendix~\ref{app:cxg} for an overview of CxG). Specifically, we focus on phrasal constructions (Cxns) because of the body of psycholinguistic research demonstrating that speakers are able to abstract over syntactic slots of these Cxns in order to interpret and produce creative and novel Cxn instantiations \cite{johnson2013evidence,Tomasello_2005}. Speakers recognize the syntactic structures of a familiar Cxn in order to interpret the meaning, despite the fact that the speaker may have never encountered that set of lexical items within the Cxn. The Exp. 1 column of Table~\ref{tab:cxnli_exp1_exp2} provides an example premise where our human annotations show that people can easily recognize the (\textsc{Resultative}) constructional semantics despite the fact that the verb (``brush'') is relatively atypical in this Cxn, which is  realized with a relatively limited set of verbs (frequently ``make'').  Additionally, speakers can balance knowledge of lexical semantics against the constructional semantics in order to distinguish Cxns that are syntactically identical but have different meanings. The Exp. 2 column of Table~\ref{tab:cxnli_exp1_exp2} provides an example premise involving a (\textsc{Depictive}) Cxn that is syntactically identical to the \textsc{Resultative}, but has a divergent meaning, as evidenced by the fact that a templatically similar hypothesis holds the opposite relation. 

In addition to providing experimentally validated explanations of human language acquisition and use, CxG is uniquely suited to evaluating whether LLMs primarily derive meaning from the composition of lexical meanings (as would be the view of Generative Grammar \cite{chomsky1995minimalist}), or if the recognition of particular syntactic structures can cue constructional meaning (this would be evidenced by strong performance on Exp. 1), or finally, if LLMs can balance both lexical meaning and constructional meaning together to recognize constructional meaning while distinguishing between syntactically identical constructions based on lexical semantics (this would be evidenced by strong performance on Exp. 2). 

\textbf{We evaluate whether or not models can generalize knowledge of highly frequent Cxns of English to creative instantiations of those Cxns with lexical items unlikely to have been encountered within that structure in pretraining data (Exp. 1), and we evaluate model ability to recognize when the lexical items are so distinct that this cues a different Cxn with a different meaning (Exp. 2).}

To investigate LLM generalization, we conduct two experiments (described in \S\ref{sec:expdesign}) leveraging the Natural Language Inference (NLI) task. We show a summary of results in Table \ref{tab:overallresults}: all models lag behind human performance in multiple experimental settings. Exp. 1 (\S\ref{sec:exp1}) shows us that some aspects of constructional semantics are ascribed adequately for success on an NLI task; however, Exp. 2 (\S\ref{sec:exp2}) shows that even state-of-the-art reasoning models like GPT-4o and GPT-o1 show a significant performance drop when the models are asked to ascribe distinct constructional semantics to syntactically identical Cxns. In combination, our results and error analysis (\S\ref{sec:error-analysis}) highlight key differences between human and model linguistic capabilities (\S\ref{sec:discussion} and \S\ref{sec:conclusions}). This study differs from previous research in significant ways highlighted by the following contributions: 

\begin{enumerate}
    \item We create a manually-validated, diagnostic evaluation dataset of 534 NLI triples testing if LLMs ascribe the appropriate semantics to phrasal Cxns.
    \item Rather than focusing on the metalinguistic task of identifying Cxns, as most prior works have done, we use the well-established NLI task to evaluate LLM `understanding' of the underlying meaning communicated by a Cxn.
    \item We test on a variety of common English phrasal Cxns instantiated by relatively unexpected words, thereby testing the ability of models to perform understanding tasks without the aid of memorization and pattern matching from large pretraining datasets.
    \item We test some of the largest models currently available, including GPT-o1 and Llama 3 70B.
\end{enumerate}
\section{Related Work}
\label{sec:relatedwork}
The evaluation of LLMs with datasets that consider pretraining data has 
largely focused on identifying and mitigating test set data leaks~\cite{balloccu-etal-2024-leak, sainz-etal-2023-nlp}. Examples of this include work by \citet{golchin2024timetravelllmstracing} and \citet{sainz2023chatgpt} who identify data contamination using prompting. Furthermore, considerable effort has been directed toward designing datasets that minimize such leakage~\cite{zhou2025lessleakbenchinvestigationdataleakage}. While this is an important concern, preventing direct data leakage does not eliminate the possibility that models can infer answers using related information from pretraining data~\cite{madabushi2025stochasticparrotingagillms}. Another relevant line of research involves counterfactual reasoning, where standard rules of the world are slightly altered. By design, counterfactuals provide an effective way to test LLMs in scenarios where pretraining data offers little advantage, and it has been shown that the performance of LLMs does in fact drop in these cases~\cite{wu-etal-2024-reasoning, lewis2024using}.  

Starting with CxGBert \citep{Tayyar_Madabushi_Romain_Divjak_Milin_2020}, there has been substantial past work on probing language models' understanding of Cxns. These works have typically focused on either a single Cxn or a handful of Cxns, like \textsc{aann} \citep{Mahowald_2023,Chronis_Mahowald_Erk_2023,misraLanguageModelsLearn2024}, \textsc{comparative-correlative} \citep{Weissweiler_Hofmann_Köksal_Schütze_2022}, \textsc{let-alone} \citep{scivetti-etal-2025-unpacking}, \textsc{npn} \citep{scivetti-schneider-2025-construction} and more schematic phrasal Cxns \citep{Li_Zhu_Thomas_Rudzicz_Xu_2022,Veenboer_Bloem_2023}. \citet{Tseng_Shih_Chen_Chou_Ku_Hsieh_2022} focus on Cxns in Taiwanese Mandarin, a notable exception to the works above which focus on English (see also \citealt{weissweiler-etal-2024-ucxn}; \citealt{bunzeck-etal-2025-construction}).  \citet{Zhou_Weissweiler_He_Schütze_Mortensen_Levin_2024} introduce NLI as a proxy task for understanding Cxns, though their results are limited to the \textsc{Causal-Excess} and related Cxns. We expand on the use of NLI to a broad set of new Cxns, and create NLI examples which utilize both entrenched (Exp. 1) and syntactically-identical (Exp. 2) Cxns. 

Additionally, recent studies show that small LMs (e.g., BabyLMs, \citealt{warstadt-etal-2023-findings}) can learn the forms of rare constructions (\citealp{misraLanguageModelsLearn2024,rozner-etal-2025-babylms}; see \citet{rozner-etal-2025-constructions} for strong formal results using RoBERTa \citep{liu2019roberta}). In contrast, we focus on generalization of constructional \textit{understanding}, which has been shown to be difficult even for LLMs in few-shot settings (\citealp{bonial-tayyar-madabushi-2024-construction-grammar,Zhou_Weissweiler_He_Schütze_Mortensen_Levin_2024}; see \citealt{mackintosh-etal-2025-evaluating} regarding impact of fine-tuning).
\textbf{To our knowledge, no prior work has explored the use of cognitive linguistic principles to generate human-generalizable datasets for assessing the generalization capabilities of LLMs.}

\section{Experimental Design}
\label{sec:expdesign}

We conduct experiments exploring two questions:

\noindent \textbf{Research Question (RQ) 1: To what extent can models generalize constructional semantics to relatively infrequent instantiations of common Cxns?} Exp. 1 uses NLI to evaluate understanding of constructional semantics in cases where high-frequency constructional templates are instantiated by words not commonly found within that Cxn. 
We select 8 Cxns that are roughly balanced across two types: \textit{argument structure Cxns} with no fixed words but clear syntactic slots (e.g., \textsc{caused-motion} in Table~\ref{tab:cxn-examples}) and  phrasal Cxns with two or more fixed words that clearly identify the Cxn (e.g., \textsc{let-alone} in Table~\ref{tab:cxn-examples}). To evaluate if the appropriate constructional semantics are associated with these Cxns, we create a novel NLI dataset, where premises are derived from corpus instances of Cxns and an understanding of constructional semantics is required to determine entailment.

\textbf{RQ 2: To what extent can models distinguish the semantics of Cxns that are syntactically identical, but have different meanings?} Exp. 2 uses NLI to evaluate abstraction of distinct constructional semantics given identical syntactic phrasings. We select five Cxns that are syntactically identical to the five argument structure Cxns of Exp. 1. We create a second set of NLI instances again using corpus instances of the five semantically distinct, but syntactically identical Cxns. 

Parallel to psycholinguistic research on analogical extension of Cxns \cite{bybee2010language}, we hypothesize that the frequency and \emph{entrenchment} of the Cxn contribute to model ability to understand the constructional semantics. There is abundant corpus linguistic data from both web and even child language indicating that the 5 argument structure Cxns tested in Exp. 1, \textsc{Causative-with, Caused-Motion, Conative, Intransitive Motion, Resultative}, are some of the earliest acquired and most frequently used Cxns in the English language \cite{hoffmann2022construction,Tomasello_2005,gries2004extending}. Although we lack CxG-annotated resources and access to pretraining data to calculate the precise frequencies of the Cxns tested in Exp. 2 (see Table~\ref{tab:cxnli_parallels}), we can safely assume that these Cxns are less frequent in the language.\footnote{We do not argue that the type frequencies of all Exp. 2 Cxns are low (e.g., \textsc{Intransitive} has high type frequency), rather that the specific sub-Cxns of Exp. 2 (e.g., \textsc{Intransitive}+"at"-\textsc{Locative}) are lower than type frequencies of argument structure Cxns of Exp. 1.}  Most importantly, while the Cxns of Exp. 1 are generally instantiated by a more limited set of verbs, the Cxns of Exp. 2, such as the \textsc{Depictive} and \textsc{Locative}, can co-occur felicitously with any verb. As a result, there is no single, \emph{entrenched variant} of the Cxns in Exp. 2 \cite{gries2004extending}. Thus, the Cxns of Exp. 1 have higher type frequency (the frequency of, for example, the \textsc{Resultative} overall) and have at least one entrenched variant with high token frequency (the frequency of the \textsc{Resultative} with ``make'').  Higher entrenchment of one variant means that there is a strong lexico-syntactic signature associated with a specific meaning of a Cxn. Lower entrenchment indicates that there is more variation in the lexico-syntactic features and more variation in how identical lexico-syntactic features are associated with several meanings.  \textbf{Thus, greater entrenchment may provide critical priors for the model to generalize constructional semantics to novel instantiations, whereas these priors may not be available for the Exp. 2 Cxns, which lack any entrenched, high token-frequency exemplar.}

\begin{table}[t]
\centering
\begin{tabular}{p{2.3cm}|p{4.7cm}}
\toprule
\textbf{Cxn Name} & \textbf{Example} \\
\midrule
Causative-With &
  \textit{Freshly ground coffee beans filled the room with a seductive, earthy aroma.} \\
\hline
Caused-Motion &
  \textit{But we also exported nickel to the United States.} \\
\hline
Comparative-Correlative &
  \textit{The more I studied, the less I understood.} \\
\hline
Conative &
  \textit{Jake sipped at the jug and didn't answer.} \\
\hline
Intransitive Motion &
  \textit{Armed troops marched to the substations and turned the power back on.} \\
\hline
Let-Alone &
  \textit{None of these arguments is notably strong, let alone conclusive.} \\
\hline
Resultative &
  \textit{He hammered the metal flat.} \\
\hline
Way-Manner &
  \textit{A middle-aged man eased his way into the room.} \\
\bottomrule
\end{tabular}
\caption{8 Cxns tested in Exp. 1, alongside examples.}
\label{tab:cxn-examples}
\end{table}

\section{Experiment 1: NLI for Constructional Semantics}
\label{sec:exp1}
\subsection{Dataset}
In Exp. 1, we leverage the  CoGS corpus \cite{bonial-tayyar-madabushi-2024-construction-grammar}, which is a collection of about 500 corpus instances, roughly balanced across 10 different Cxns. CoGs consists of carefully curated Cxns chosen for their broad coverage of the basic phrasal Cxns of English \cite{hoffmann2022construction}. The Cxn types collectively represent a significant portion of English usage and provide an effective basis for evaluating LLMs on high-frequency Cxns (such as the \textsc{Caused-Motion}), where instantiated with creative words. 

Broadly, CoGS consists of Cxns of two types: argument structure Cxns, which involve no fixed word forms but have been shown to be the most common Cxns of English as well as other languages \cite{Goldberg_1995}; and phrasal Cxns with multiple fixed words. The latter Cxns are more easily recognizable to LLMs given fixed words cueing that Cxn \cite{bonial2024constructing}. We construct our datasets with 8 of the 10 Cxns in CoGS, shown in Table \ref{tab:cxn-examples}. 


Overall, our process for creating the constructional NLI dataset can be summarized as:
\begin{enumerate}[itemsep=0.1em, parsep=0pt]
    \item Extract corpus Cxn examples from CoGS.
    \item Create general templates for NLI hypotheses for each Cxn type.
    \item Generate hypotheses for each example Cxn given the corresponding templates.
    \item Manually validate the resulting dataset.
\end{enumerate}
We explain this process through the following example, beginning with the premise: \textit{``He hammered the metal flat.''} This is a \textsc{resultative} Cxn, which has the meaning of an action causing a change in state. In this case, the action verb \textit{hammered} leads to \textit{the metal} to have a resulting state of \textit{flat}. Regarding the syntactic form of a phrasal Cxn, we can use a constructional template to define the syntactic nature of the slots that are filled by a given Cxn. A general template for the \textsc{resultative} is shown in Example \ref{ex:res}, and its application to the above sentence in Example \ref{ex:res2}.

\ex .\label{ex:res} [SBJ\textsubscript{1} [V\textsubscript{2} OBJ\textsubscript{3} ADJ\textsubscript{4} ]\textsubscript{VP}]\textsubscript{5} 

\ex .\label{ex:res2} [\textit{He}\textsubscript{1} [\textit{hammered}\textsubscript{2} \textit{the metal}\textsubscript{3} \textit{flat}\textsubscript{4} ]\textsubscript{VP}]\textsubscript{5} 

Given constructional examples like those in Example \ref{ex:res2}, our goal is to produce NLI tuples that consistently target the Cxn's meaning. We do this by manipulating the slots in the Cxn templatically to produce hypotheses with consistent relations to the premise. For instance, consider the hypothesis \textit{``The hammering did not cause the metal to become flat.''} This is clearly a contradiction to the above premise and is directly in conflict with the meaning of the Cxn. We can produce this hypothesis, and similar contradicted hypotheses, by following the template in Example \ref{ex:res3}, instantiated with our current example in Example \ref{ex:res4}. 

\ex .\label{ex:res3} [\textsc{The} [V]\textsubscript{2}-\textsc{ing} \textsc{did not cause} [OBJ]\textsubscript{3} \textsc{to become} [ADJ]\textsubscript{4}].

\ex .\label{ex:res4} [\textsc{The} [\textit{hammer}]\textsubscript{2}-\textsc{ing} \textsc{did not cause} [\textit{the metal}]\textsubscript{3} \textsc{to become} [\textit{flat}]\textsubscript{4}].

Thus, for each Cxn, we create a template to generate NLI hypotheses that target constructional meaning.
\footnote{To ensure that the templates did not bias the models in some way towards the appropriate NLI relation, we produced free-form NLI triples for the same premises in a separate research effort; we found that model performance improves on the freeform hypotheses over our templated hypotheses \cite{bonial2025form2function}. This indicates that the templates do not seem to cue the model to the correct relation.} 
See Appendix \ref{sec:appendix_NLI_Examples} for the NLI templates for each Cxn (Table~\ref{tab:templates}), along with examples (Table~\ref{tab:nli-examples_exp1}). 

\textbf{Manual Verification} We manually validate every test instance of our dataset with double or triple annotation and measure human agreement (ranging from 78-90\%) to ensure the robustness of our claim that people are able interpret the specific Cxn instances that we present to the LLMs.  Once the dataset was created, a second and sometimes third author annotated the relations, and hypotheses were amended until achieving an Inter-Annotator Agreement (IAA) of 90\%. Thus, if we take the original author's assigned relation to be the gold standard, then native speaker accuracy on the NLI task is 90\%. The final Exp. 1 ``CxNLI'' dataset totals 435 triples.  Descriptive statistics for this dataset along with all other datasets can be found in Appendix \ref{sec:appendix_stats} (Table~\ref{tab:datasetstats}). 
\subsection{Formalism and Task Design}
We define a Cxn, $C$, to be a pairing of a \textbf{constructional schema} (form), $\mathcal{T}_C$, and a \textbf{semantic interpretation} (meaning), $M(C)$. For example, for the \textsc{Resultative} Cxn has the \textbf{schema ($\mathcal{T}_C$)} \texttt{NP V NP ADJ} and \textbf{Meaning ($M(C)$)} `The action of the Verb causes the Object to enter the state described by the Adjective.'

Our method of evaluating models' ability to `understand' this Cxn  begins with a premise sentence, $p$, that is an instance of a given Cxn $C$ ($p \in C$). We then generate a hypothesis, $h$, by applying a pre-defined \textbf{hypothesis template}, $\mathcal{H}_{C,L}$. This template is a function that takes the premise $p$ as input, extracts its relevant components, and generates a new sentence $h$. The template is designed to probe the Cxn's core meaning, $M(C)$, in a way that produces a predictable NLI label, $L \in \{\text{Entailment, Contradiction}\}$. Therefore, the hypothesis is generated as: $h = \mathcal{H}_{C,L}(p)$. For example, to generate a contradiction ($L=\text{Contradiction}$) for a \textsc{Resultative} premise:
\begin{itemize}
  \setlength\itemsep{-0.1em}
    \item \textbf{Premise ($p$):} ``He hammered the metal flat.''
    \item \textbf{Hypothesis ($h$):} $h = \mathcal{H}_{\textsc{Resultative}, \text{Contradiction}}(p) \Rightarrow$ ``The hammering did not cause the metal to become flat.''
    \item \textbf{Resulting Tuple:} $\langle p, h, \text{Contradiction} \rangle$.
\end{itemize}
The goal of this experiment is to assess if models can correctly classify these NLI tuples. High accuracy on this task indicates that a model has learned the fundamental association between a syntactic schema $\mathcal{T}_C$ and its meaning $M(C)$, even when instantiated with creative or atypical words.
\subsection{Empirical Evaluation and Analysis}
We test three OpenAI models on our constructional NLI dataset: GPT-4o-2024-05-13 and GPT-3.5-turbo-0125, as well as o1-preview-2024-09-12.\footnote{\url{https://platform.openai.com/docs/models}} We also test two Llama models: Llama-3-8B-instruct and Llama-3-70B-instruct. These models were chosen for their large sizes, which make them illustrative examples of the capabilities of state-of-the-art LLMs in general. We test 3 main scenarios: \textit{zero-shot}, \textit{in-context learning} with examples randomly selected from Stanford NLI (SNLI, \citealt{bowman2015large}), and \textit{in-context learning} with Constructional NLI (CxNLI, our dataset).\footnote{We experiment with a variety of prompt formats and report results for the best performing prompts; details can be found in Appendix \ref{sec:appendix_prompt_variation}. We also perform Chain-of-Thought (CoT) experiments \citep{Wei_Wang_Schuurmans_Bosma_Ichter_Xia_Chi_Le_Zhou_2022} in each of these scenarios; it does not lead to performance gains: see Appendix \ref{sec:appendix_cot}.} A summary of our results for Exp. 1 are reported in Table \ref{tab:exp1scores}.

\begin{table}[t]
\centering
\begin{tabular}{p{1cm}p{1cm}p{0.5cm}p{0.5cm}p{0.5cm}p{0.5cm}p{0.5cm}}
\hline
\textbf{Setting} & \textbf{IC} & \multicolumn{5}{c}{\textbf{Accuracy}}\\
&  \textbf{Data} & \multicolumn{3}{c}{\textbf{GPT}} & \multicolumn{2}{c}{\textbf{Llama 3}}\\
 & & 3.5 & 4o &o1* & 8B & 70B\\
\hline

0-shot & None & 0.6 & 0.88 & \textbf{0.90}& 0.59 & 0.74 \\
1-shot & CxNLI & 0.69 & 0.9 & - & 0.65 & 0.84 \\
3-shot & CxNLI & \textbf{0.79} & \textbf{0.96} & \textbf{0.90} & \textbf{0.73} & \textbf{0.91} \\
1-shot & SNLI & 0.58 & 0.86 & - & 0.59 & 0.75 \\
3-shot & SNLI & 0.59 & 0.86 & 0.89& 0.58 & 0.74 \\
\end{tabular}%
\vspace{-0.1in}
\caption{Results for Exp. 1 - Evaluation on our CxNLI dataset. ``IC Data" refers to the type of data used as in-context examples.*GPT-o1 is only tested in zero-shot and three-shot settings on a subset of the overall data due to API costs. }
\label{tab:exp1scores}
\end{table}

Overall, we see that performance is high even in the zero-shot setting for GPT-4o and GPT-o1. We also observe that GPT-4o and Llama 3 70B consistently perform better than their smaller model counterparts GPT-3.5 and Llama 3 8B. Adding examples of Cxns for in-context learning boosts performance, while additional SNLI examples do not boost performance. This is especially true for GPT-3.5 and Llama 3 8B, which benefit substantially more from in-context learning from CxNLI. This reliance on in-context learning indicates that our datasets test a different axis of semantic knowledge than more general datasets like SNLI. 
\section{Experiment 2: NLI for Distinguishing Syntactically-Identical Cxns}
\label{sec:exp2}
\subsection{Dataset}
In Exp. 1, we show that the models perform impressively on an NLI task that specifically targets constructional semantics. The Cxns tested are common to the English language, and the templates we generate target aspects of meaning that are highly salient for the Cxn. 
In Exp. 2, we test whether models can generalize the appropriate constructional semantics for syntactically identical phrasal Cxns that should be ascribed distinct semantics. This enables us to determine if models have a robust capability to attribute and understand constructional semantics, or if this capability might be limited to the more entrenched phrasal Cxns of the language that were tested in Exp. 1.  

\begin{table*}[t]
\centering
\label{tab:exp1-exp2}
\resizebox{\textwidth}{!}{
\begin{tabular}{llll}
\multicolumn{2}{l}{\textbf{Exp. 1}} & \multicolumn{2}{l}{\textbf{Exp. 2}} \\
\toprule
\textbf{Cxn} & \textbf{Argument Structure Cxn} & \textbf{Syntactically-Identical New Cxn} & \textbf{Cxn} \\
\midrule
Resultative         & I brushed my hair smooth.                    & A famous emperor buried scholars alive.  & Depictive           \\
Conative            & Marco grabbed at the ladder railing.         & Other units exploded at this complex.    & Intransitive+"at"   \\
Caused-motion       & We exported nickel to the United States      & I introduced her to my boss.             & DitransitiveV+NP+PP \\
Intransitive-motion & The 23 scrambled to the rear of the sub.     & They're listening to the same podcast.   & Intransitive+"to"   \\
Causative-with      & Samsung flooded the market with advertising. & I use a mouse with my left hand.         & Transitive+"with"   \\
\bottomrule
\end{tabular}
}
\caption{Example CxNLI premises from Exp. 1 and 2, illustrating syntactically identical, semantically distinct Cxns.}
\label{tab:cxnli_parallels}
\end{table*}

Thus, for the 5 argument structure Cxns of our 8 Cxns, we add test instances which share a surface syntax with our Cxns, but convey a different meaning.\footnote{We test 5 of 8 Cxns from Exp. 1 because 3 Cxns lack syntactically identical counterparts with distinct meanings.} Table~\ref{tab:cxnli_parallels} provides examples of the original Cxns used in Exp. 1 and parallel, syntactically identical Cxns tested in Exp. 2.  Consider the following: 

\ex. \label{ex:adv1} He hammered the metal flat. \textit{(resultative)}

\ex. \label{ex:adv2} I bought the apples fresh. \textit{(depictive)}

In the above two examples, the syntactic forms are identical: both have a subject pronoun, a verb, an object noun phrase followed by an adjective. However, the two Cxns convey different meanings: In Example \ref{ex:adv1} the adjective is the \textit{result} of the action of the verb, whereas in Example \ref{ex:adv2}, the adjective is the state of the noun \textit{during} the action of the verb, but it is \textit{not} the resulting state of the action. This difference in meaning is associated with two different Cxns, specifically the \textsc{resultative} and the \textsc{depictive}. We can tease apart this difference in meaning by leveraging our template-based hypotheses from our CxNLI dataset in \ref{sec:exp1}. Consider the following templatic hypotheses:

\ex. \label{ex:adv3} [\textsc{The} [\textit{hammer}]\textsubscript{2}-\textsc{ing} \textsc{caused} [\textit{the metal}]\textsubscript{3} \textsc{to become} [\textit{flat}]\textsubscript{4}]. \textit{(entailment)}

\ex. \label{ex:adv4}  [\textsc{My} [buying]\textsubscript{2}-\textsc{ing} \textsc{caused} [\textit{the apples}]\textsubscript{3} \textsc{to become} [\textit{fresh}]\textsubscript{4}]. \textit{(contradiction)}

As we can see in Examples \ref{ex:adv3} and \ref{ex:adv4}, templatically generating hypotheses for these two examples leads to different relations to the premises. The Cxns we use for this dataset are \textsc{Intransitive + PP\textsubscript{At}}, \textsc{Intransitive + PP\textsubscript{to}}, \textsc{Ditransitive} with NP, PP complements, \textsc{Transitive + PP\textsubscript{with}}, and the \textsc{Depictive}.\footnote{More examples of each of these Cxns along with NLI tuples are shown in Appendix \ref{sec:appendix_NLI_Examples}, Table \ref{tab:cnli-9-cxn-adv-examples}.} 

\textbf{Manual Verification} After the final version of this dataset was created, a second author evaluated the dataset, achieving an IAA of 83\% with the original judgments. The final Exp. 2 ``CxNLI-Distinction'' dataset totals 99 NLI triples. 
\vspace{-0.06in}
\subsection{Formalism and Task Design}
\vspace{-0.06in}
Let $C$ be the target Cxn (e.g., \textsc{Resultative}) and $C'$ be the syntactically identical distractor Cxn (e.g., \textsc{Depictive}). The Cxns are selected such that the entrenchment of the distractor is lower than that of the target: $ Entr(C') < Entr(C)$
While both Cxns are realized via the same \textbf{constructional schema}, meaning their syntactic templates (e.g., \texttt{NP V NP ADJ}) are identical ($\mathcal{T}_C = \mathcal{T}_{C'} = \mathcal{T}$), their underlying semantic interpretations are distinct, $M(C) \neq M(C')$. 

Our evaluation focuses on premises that are instances of the distractor Cxn $C'$. As an illustration, consider one such premise, $p'$:
\begin{quote}
    Schema: $\mathcal{T}_{C'} = \texttt{NP V NP ADJ}$ \\
    Example: $p' \in C' \Rightarrow$ ``A famous emperor buried China’s scholars alive...''
\end{quote}
The core of the experiment is to generate \textbf{two different hypotheses} for this single premise, each designed to probe for the meaning of either the (correct) distractor ($M(C')$) or (incorrect) target Cxn ($M(C)$). 
\vspace{-0.06in}
\begin{enumerate}
    \item \textbf{An NLI tuple probing the correct (\textsc{Depictive}) meaning:} 
    \begin{itemize}
        \item $h_1 = \mathcal{H}_{C', \text{Entailment}}(\mathbf{p'}) \Rightarrow$ ``China's scholars were fully alive before being buried."
        \item This creates the NLI tuple: $\langle p', h_1, \text{Entailment} \rangle$.
    \end{itemize}

    \item \textbf{An NLI tuple probing the incorrect (\textsc{Resultative}) meaning:} 
    \begin{itemize}
        \item $h_2 = \mathcal{H}_{C, \text{Contradiction}}(\mathbf{p'}) \Rightarrow$ ``Burying caused the scholars to become alive.''
        \item This creates the NLI tuple: $\langle p', h_2, \text{Contradiction} \rangle$.
    \end{itemize}
\end{enumerate}
\vspace{-0.06in}
Crucially, this experimental design assesses if models can avoid over-generalizing the meaning of the more entrenched target Cxn, $M(C)$, to a less entrenched distractor Cxn, $C'$, that shares the same syntactic schema ($\mathcal{T}_{C'} = \mathcal{T}_C$) but has a distinct semantic interpretation ($M(C') \neq M(C)$). By measuring overall accuracy, we test a model's ability to reject the semantics of the highly entrenched Cxn ($M(C)$) when presented with a premise from the less entrenched distractor, while simultaneously accepting the correct meaning ($M(C')$). We use performance on the more straightforward instances in Exp. 1 as a baseline; therefore, a significant drop in accuracy on this second task indicates a failure to distinguish Cxns based on their subtle semantic cues, highlighting a key difference from human generalization. 

\vspace{-0.06in}
\subsection{Empirical Evaluation and Analysis}
\vspace{-0.06in}
We utilize this new Exp. 2 dataset as an evaluation dataset to test if LLMs can perform NLI successfully on the new phrasal Cxn examples, where performance requires distinguishing the unique semantics of the syntactically identical Cxn. As we can see in Table \ref{tab:exp2bscores}, performance is significantly lower than our results from Exp. 1 in almost every prompt setting and across all models. The difference in performance is stark.  
While these examples are also slightly 
more difficult for humans, the ceiling of human performance (IAA 83\%) is well above current LLM performance, even for GPT-4o, GPT-o1 and Llama 3 70B. Again we see a large difference between the GPT-4o and the smaller models, and also see that GPT-o1 performs worse than GPT-4o. 

We take these results as evidence that the ability of models to abstract over the same syntactic slots and assign the appropriate semantics is limited to the more entrenched Cxns of English, of which our original Exp. 1 dataset consisted. By shifting to phrasal Cxns that are syntactically identical but potentially less entrenched, we are essentially testing whether or not models can abstract over less data to arrive at a less statistically likely semantic interpretation. This also highlights that the abstraction process in people clearly goes beyond the syntactic character of the slots alone---people are able to balance their knowledge of lexical semantics with their knowledge of constructional semantics to arrive at the most pragmatically likely interpretation.  People's lexical awareness is also imbued with physical world knowledge; thus, people are abstracting over a distinct set of information than what LLMs have access to. 

\begin{table}[h]
\centering
\begin{tabular}{p{1cm}p{1cm}p{0.5cm}p{0.5cm}p{0.5cm}p{0.5cm}p{0.5cm}}
\hline
\textbf{Setting} & \textbf{IC} & \multicolumn{5}{c}{\textbf{Accuracy}}\\
&  \textbf{Data} & \multicolumn{3}{c}{\textbf{GPT}} & \multicolumn{2}{c}{\textbf{Llama 3}}\\
 & & 3.5 & 4o &o1* & 8B & 70B\\
\hline

0-shot & None & 0.26 & \textbf{0.58} & \textbf{0.46}& 0.38 & 0.52\\
1-shot & CxNLI & 0.26 &  0.53 & -& 0.29 &  \textbf{0.60} \\
3-shot & CxNLI & \textbf{0.31} & 0.57 & 0.45& 0.35 & 0.50 \\
1-shot & SNLI & 0.3 & 0.55&-& 0.37 & 0.54\\
3-shot & SNLI & 0.28 & 0.53 & \textbf{0.46}& \textbf{0.39} & 0.57\\
\end{tabular}%
\vspace{-0.1in}
\caption{Results for Exp. 2, testing on CxNLI-Distinction. "IC Data" refers to the type of data used in the in-context examples. *GPT-o1 is only tested in zero-shot and three-shot settings due to limited resources.}
\label{tab:exp2bscores}
\end{table}

\subsection{Statistical Analysis}
We tested whether performance on CxNLI-Distinction (Exp. 2) is worse than on CxNLI (Exp. 1) for humans and for each model.  Specifically, we conducted a Bayesian A/B test to quantify the evidence for a performance drop on the more challenging CxNLI-Distinction (Exp. 2) dataset. This analysis was performed separately for human participants and three different models: LLaMA, GPT-4o, and GPT-o1.  We assigned a strong prior belief of 99\% that performance on the CxNLI-Distinction dataset would be equal to the CxNLI (Exp. 1) dataset and a 1\% probability of lower performance.

\textbf{Despite encoding strong priors favoring equal performance, our data consistently point to a performance drop.} The analysis, based on 10,000 posterior samples, gives a Bayes Factor (BF) for each group. Using the standard interpretation where a Bayes Factor over 10 constitutes strong evidence, we find that for humans, the data are 4.3 times more likely under the hypothesis that performance is worse (BF = 4.31). Given that a BF of 10 constitutes strong evidence, this is only weak evidence that the performance is worse. In contrast, \textbf{for LLaMA the evidence is very strong}, with the data being over 2,600 times more likely under this hypothesis (BF\textsubscript{10} = 2,684). Furthermore, \textbf{for GPT-4o and o1 the evidence is extreme} with the BFs being 3.6 x 10\textsuperscript{8} and 8.2 x 10\textsuperscript{16} respectively.

Thus, the evidence overwhelmingly supports the hypothesis that the CxNLI-Distinction (Exp. 2) set is significantly more difficult for LLMs than for humans. Additionally, this demonstrates that our CxNLI-Distinction dataset size is large enough to conclude that performance is decisively worse on this dataset, which requires distinguishing between syntactically identical constructions.

\section{Error Analysis}
\label{sec:error-analysis}
In Exp \hyperref[sec:exp1]{1}, we describe our CxNLI experiments, which find that GPT-4o, GPT-o1 and Llama 3 70B are extremely proficient at CxNLI while GPT-3.5 and Llama 3 8B lag behind substantially. In Exp \hyperref[sec:exp2]{2}, we show that all our tested models do not perform well when tested on our CxNLI-Distinction dataset with five additional, syntactically identical Cxns. For example, though GPT-4o's performance on the CxNLI \textsc{resultative} is near perfect, it struggles to demonstrate understanding of the syntactically identical \textsc{depictive:} 
\ex. \label{ex:adv-error} \textbf{Premise:} \textit{I bought the apples fresh.} \\
\textbf{Hypothesis:} \textit{The apples were completely fresh before I bought them.} \\
\textbf{Correct Response:} Entailment\\
\textbf{Model Response:} Contradiction

Here, we investigate if some Cxns are harder for LLMs than others. Among Cxns from Exp. 1, \textsc{let-alone} and \textsc{comparative-correlative} are the weakest Cxns for GPT-4o, though it is strong across the board with a minimum accuracy of 88\%. GPT-3.5 is much more variable by Cxn, with a maximum accuracy of 92\% for the \textsc{conative} and a minimum of 67\% for \textsc{let-alone}. We show an example of GPT-4o misunderstanding the scale of \textsc{let-alone} in Example \ref{ex:cnli-error}.

\ex. \label{ex:cnli-error} \textbf{Premise:} \textit{Beecher's reputation as a preacher, let alone as a Man of God, was not universally accepted.} \\
\textbf{Hypothesis:} \textit{Beecher's reputation as a Man of God was easier to accept than his reputation as a preacher.} \\
\textbf{Correct Response:} Contradiction\\
\textbf{Model Response:} Entailment 

\begin{figure}[t]
    \centering
    \includegraphics[width=0.9\linewidth]{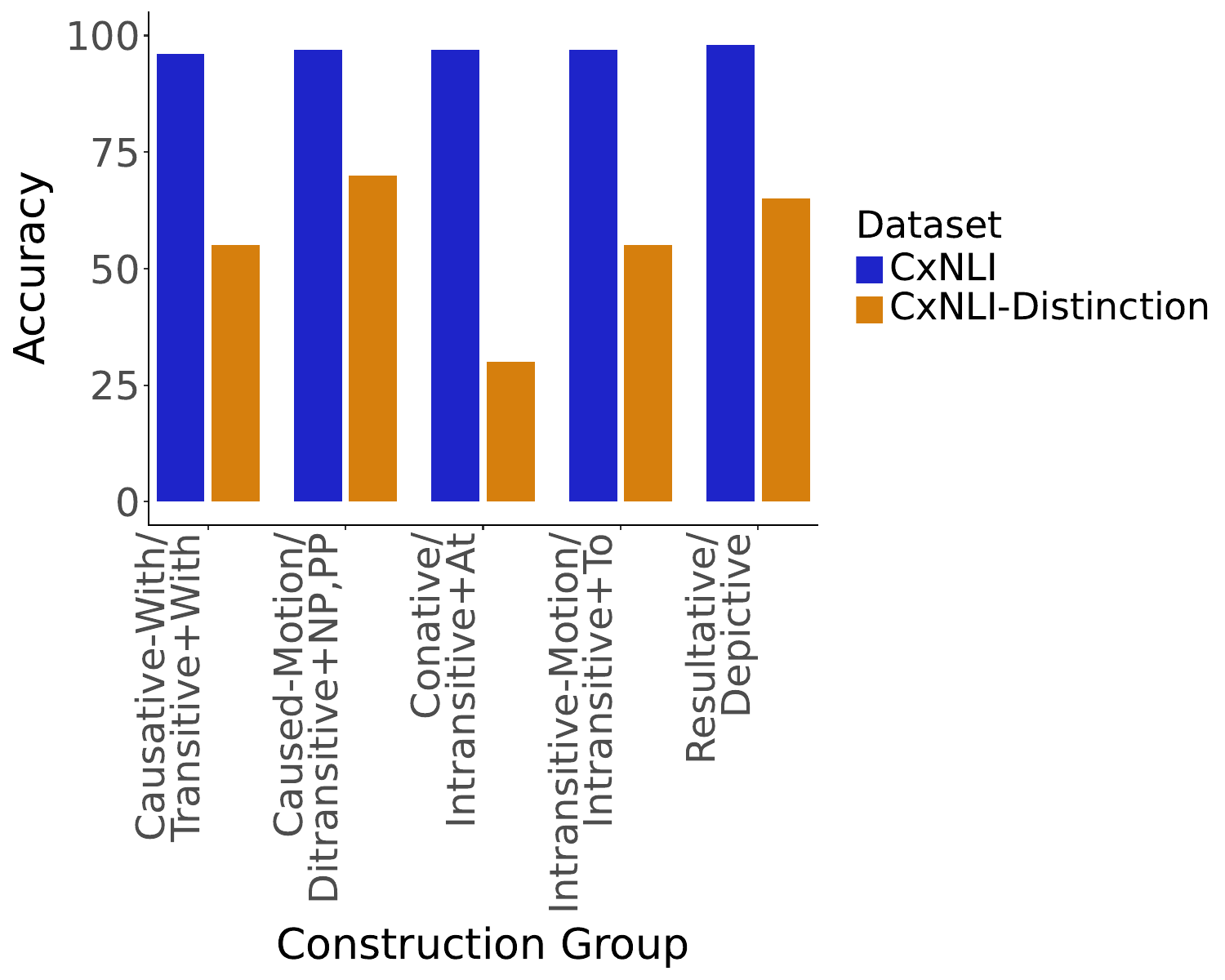}
    \vspace{-0.1in}
    \caption{Constructional NLI Accuracy broken down by Cxn (for best prompts). Accuracy drops substantially for the Cxns in Exp. 2 relative to those in Exp. 1.}
    \label{fig:cnliscores}
\end{figure}

In Figure \ref{fig:cnliscores} we report the accuracy by Cxn in our Exp. 1 NLI and Exp. 2 NLI datasets.\footnote{Accuracies are from the highest performing prompt. We only visualize GPT-4o for visual clarity, though trends are similar across models.} We see performance is  lower for all Cxns we test in Exp. 2. This provides evidence supporting our hypothesis, outlined in \S\ref{sec:expdesign}, that the entrenchment of a Cxn contributes to model ability to understand the constructional semantics. 
 
\section{Discussion}
\label{sec:discussion}
Our results show a clear discrepancy in the ability of LLMs to process constructional meaning. While models demonstrated a surprisingly high capacity to interpret familiar constructions even with novel lexical fillers (Exp. 1), their performance dropped significantly when required to distinguish between syntactically identical constructions that carry different meanings (Exp. 2). 
This discrepancy highlights a failure in how models generalize from frequent patterns to more nuanced, creative uses of language and has significant implications to the development and use of LLMs

\textbf{Data, Bias, and the Limits of Scale} Our Exp. 2 findings challenge the prevailing ``more data is better'' paradigm~\cite{DBLP:journals/corr/abs-2001-08361}. The models' bias towards the most frequent constructional meaning suggests that simply scaling web-text data may reinforce these errors. 
This implies a need for new data strategies, such as up-sampling rare-but-important structures or using targeted~\cite{ye2025data}, adversarial fine-tuning~\cite{NEURIPS2021_22b1f2e0} to correct the biases of the base model.

\textbf{The Need for Diagnostic Evaluation} The significant difference in performance between our two experiments shows how broad-coverage benchmarks can overestimate a model's true linguistic competence. Our work demonstrates the importance of contrastive, diagnostic benchmarks to test specific, theoretically-grounded phenomena. 

\textbf{A Failure of Causal Reasoning and Its Safety Implications} The model's inability to distinguish a \textsc{Depictive} from a \textsc{Resultative} construction is a failure to reason about causality. \emph{This is not a niche linguistic error;} it has direct implications for AI safety as misunderstanding the difference between a co-occurring state and a caused outcome could lead to catastrophic errors.

\textbf{Architectural Limitations and Future Directions} \emph{The systematic nature of this error across models suggests the issue may be rooted in the architecture of these models.} Therefore, the path forward may benefit from the use of novel and hybrid architectures, such as the incorporation of constructional resources or the addition of long term memory~\cite{wang2023augmenting}.

\section{Conclusions and Future Work}
\label{sec:conclusions}
We have shown where even the latest models do not demonstrate a functional understanding of Cxns: although models can generalize the semantics of entrenched Cxns to creative  instantiations, the same models cannot robustly distinguish between syntactically identical Cxns with distinct semantic interpretations. While GPT-4o, GPT-o1 and Llama 3 70B do perform quite impressively on our original constructional NLI task, they fail at the NLI scenario requiring constructional distinction, which requires generalization of the appropriate constructional semantics to syntactically-identical Cxns. Also, we see that GPT-4o substantially outperforms GPT-3.5 in all settings, and in-context learning is especially crucial for GPT-3.5. Overall, these experiments show that the constructional awareness of GPT-4o, GPT-o1 and Llama 3 70B are far more robust than that of GPT-3.5 and Llama 3 8B, but their ability to generalize constructional meaning to both novel instantiations and distinct Cxns still lags substantially behind that of humans. 

Thus, our targeted series of experiments demonstrate that LLMs do process constructional semantics up to a point, yet our datasets and experiments reveal the breaking point of understanding---where speakers are able to recognize the appropriate constructional semantics despite both novel instantiations and despite the fact that there are multiple, syntactically identical Cxns that could be candidates for interpreting the phrase at hand. Overall, we find that CxG serves as a valuable theoretical lens for probing the functional language understanding of LLMs with a methodology that tests for linguistic generalization beyond memorization and dependency on pretraining priors and comparing this with human linguistic knowledge. 
Greater contributions to resources such as corpora of Cxns will facilitate empirical data on which constructional understanding can be evaluated with more detail.

\section*{Limitations}
This work is limited in that we only evaluate our methods on English. More work is needed on the targeted evaluation of LLM performance using Cxn information in non-English settings. Our tasks are only one possible method for investigating LLM understanding of Cxns. Expanding research to include complementary methodologies will be necessary to build a complete picture of LLM knowledge in relation to CxG. This work can also be extended beyond the 8 Cxns that we use to generate our dataset, although these were selected for the extensive coverage of the English language. Also, while we consciously choose to create a smaller, more carefully curated dataset that also allows for careful expert manual evaluation, there is scope to increase the size of our dataset, which we leave to future work.  

\section*{Ethics}

LLMs are extremely expensive to train and run. The compute costs associated with LLMs have a nontrivial environmental impact which should not be ignored. Furthermore, due to their large-scale training data, they can reflect and propagate harmful social biases in their responses if they are not properly aligned and moderated. Furthermore, there is risk of LLMs having a negative societal impact if their widespread deployment is done without proper consideration for the lives of people. While there are risks in the use and proliferation of LLMs in general, we do not believe this work incurs any specific additional risks. Despite the overall risks and dangers, we believe this research is worthwhile in order to better understand the language systems of LLMs and compare and contrast LLM language understanding with that of humans. We honor the code of ethics. 

\section*{Acknowledgments}
Melissa Torgbi is supported by the UKRI Centre for Doctoral Training in Accountable, Responsible, and Transparent AI [EP/S023437/1] of the University of Bath.

\bibliography{custom}

\appendix

\section{Construction Grammars}
\label{app:cxg}
CxG has particular explanatory power with respect to phrasal Cxns, such as the \textsc{resultative} Cxn: ``The jackhammer pounded us deaf.'' Generative linguistic theories (e.g., \citet{chomsky2014minimalist}) would generally analyze a transitive sentence with one verbal head (``pounded'') that licenses the arguments of the sentence. ``Pounded'' generally licenses an agent subject (here, ``jackhammer'') and potentially a patient direct object.  Generative approaches argue that this information about the verb is memorized and stored in the lexicon, while combinatory rules of how to put lexical items together are stored in a separate syntax module of language processing. However, unless a special grammatical rule or sense of the verb is postulated, there is nothing to explain why ``us'' is not the thing pounded here, or what licenses the adjective ``deaf.'' Nonetheless, native speakers have no problem recognizing the special formal and semantic properties of this Cxn, namely that it entails a pounding event that causes a change in state of ``us'' resulting in the state of ``deaf.'' 

In contrast to Generative linguistic theories (e.g., \citet{chomsky2014minimalist}), CxG posits that speakers acquire and store Cxns, which notably account for not only the semantic properties of the unit but also the formal syntactic properties. Cxns at all levels of language are learned through language usage; thus, in lieu of grammatical rules accounting for the grammaticality of a particular structure, frequency plays an important role in what `sounds right' or is grammatical to a speaker. In the CxG account, speakers acquire single-word Cxns that they are frequently exposed to (e.g., ``milk''), but then generalize from that to recognize how an acquired holophrastic Cxn falls into slots of larger, more complex Cxns (e.g., ``want milk,'' ``baby want milk,'' ``want up,'' ``I want to go''). In this process of generalization, speakers build up a taxonomically organized set of the related Cxns of their language, or a \textit{constructicon}. Speakers use domain-general cognitive processes to incrementally generalize over such frequent Cxns to novel usages, arriving at the ability to interpret even rare and previously unseen instances of a Cxn.  



\section{Scientific Artifacts and Descriptive Statistics for All Datasets}
\label{sec:appendix_stats}

We use the following scientific artifacts: COCA \citep{Davies_2010}, EnCOW \citep{Schäfer_Bildhauer_2012,Schäfer_2015}, the CoGS dataset \citep{bonial-tayyar-madabushi-2024-construction-grammar}, and the OpenAI API, in addition to our created datasets. The COCA corpus contains 8 genres: Academic, Blog, Fiction, Magazine, News, Spoken, TV, and Web. It is intended to capture American English, though there is no guarantee that it does not also include some other varieties. Demographic information about the creators of the texts in the corpus is not always available given its scale. EnCOW is a large scale corpus of English text from the web. As such, the demographic information that it captures is not completely clear. All datasets besides COCA and EnCOW are open-source under a Creative Commons license. The institutions of the authors have valid licenses for COCA and EnCOW, permitting their use in academic settings. We use the artifacts as intended. 
Overall, we construct and experiment on 2 datasets: CxNLI (Exp. 1) and CxNLI-Distinction (Exp. 2). The sizes and IAA agreement for the final datasets is reported in Table \ref{tab:datasetstats}. All of our datasets are exclusively in English.

\begin{table}[!h]
  \centering
  \small
  \begin{adjustbox}{max width=\columnwidth}
    \begin{tabular}{@{}l r r r@{}}
      \toprule
      \textbf{Dataset} & \textbf{N} & \textbf{Tokens} & \textbf{IAA} \\
      \midrule
      CxNLI (Exp. 1) & 435 & 15,144 & 90\% \\
      CxNLI-Distinction (Exp. 2) & 99 & 3,202 & 83\% \\
      \bottomrule
    \end{tabular}
  \end{adjustbox}
  \caption{Descriptive statistics for each dataset. N is the number of unique examples.}
  \label{tab:datasetstats}
\end{table}


\subsection{CxNLI (Exp. 1) Data Sources}
\label{sec:appendix_genvstemp}



The templatic dataset is constructed with real-world corpus data of Cxns, primarily coming from the CoGS dataset \citep{bonial-tayyar-madabushi-2024-construction-grammar}, with supplementary data coming from Corpus of Contemporary American English (COCA, \citealt{Davies_2010}), and the English Corpus from the Web, or EnCOW \citep{Schäfer_Bildhauer_2012,Schäfer_2015}. 
Within this dataset, each premise includes one of 8 total Cxns: the \textsc{comparative-correlative} Cxn, the \textsc{let-alone} Cxn, the \textsc{way-manner} Cxn, the \textsc{causative-with} Cxn, the \textsc{conative} Cxn, the \textsc{resultative} Cxn, the \textsc{caused-motion} Cxn, and the \textsc{intransitive-motion} Cxn. We include a roughly balanced sample of each Cxn, with all premises taken from corpus data. These Cxns cover a wide range of \textit{schematicity} meaning that they have different levels of lexicalization/abstractness. 

\section{Annotator Information}

All datasets are annotated by co-authors of this paper. 2 annotators identify as men, 4 annotators identify as women. Of our annotators, 3 have a graduate degree in linguistics while 3 do not. At least one linguistic expert and one non-expert annotate each dataset. Because the source of our datasets is web corpora, there is some risk of offensive or hateful content. During annotation, annotators were asked to remove any hateful or offensive content as well as personal identifying information. The annotation task was given to annotators in a spreadsheet. The instructions provided are detailed in Appendix \ref{appendix:nli-guidelines}. 

\section{Example NLI Tuples}
\label{sec:appendix_NLI_Examples}
Our templates for generating hypotheses across Cxn types is shown in Table~\ref{tab:templates}. In Table ~\ref{tab:nli-examples_exp1}, we show examples of our constructional NLI datasets from Exp. 1. We show examples of the distinction-requiring NLI examples from Exp. 2, alongside examples of our new constructions for Exp. 2, in Table ~\ref{tab:cnli-9-cxn-adv-examples}.

\begin{table*}[ht]
\centering
\resizebox{\textwidth}{!}{%
\begin{tabular}{p{3.4cm}|p{7.1cm}|p{7.1cm}}
\toprule
\textbf{Cxn Name} & \textbf{Cxn Template} & \textbf{NLI Hypothesis Template} \\ \midrule
\hline
\textbf{Causative-With} &
 [SBJ\textsubscript{1} [V\textsubscript{2} OBJ\textsubscript{3} \textit{with}-PP\textsubscript{4} ]\textsubscript{VP}]\textsubscript{5} & SBJ\textsubscript{1} \textit{did not cause} OBJ\textsubscript{3} \textit{to contain} OBJ-of-PP\textsubscript{4}\\ \hline
\textbf{Caused-Motion} & 
 [SBJ\textsubscript{1} [V\textsubscript{2} OBJ\textsubscript{3} PP\textsubscript{4} ]\textsubscript{VP}]\textsubscript{5} & OBJ\textsubscript{3} \textit{did not change locations}. \\
  \hline
\textbf{Comparative-Correlative} &
 [[\textit{the}\textsubscript{1} [Comparative Phrase]\textsubscript{2} REST-CLAUSE\textsubscript{3}]\textsubscript{C1} [[\textit{the}\textsubscript{4} [Comparative Phrase]\textsubscript{5} REST-CLAUSE\textsubscript{6}]\textsubscript{C2}]\textsubscript{7}& \textit{The amount} [Comparative Phrase]\textsubscript{2} \textit{is positively/negatively correlated with the amount} [Comparative Phrase]\textsubscript{5} \\ \hline
\textbf{Conative} & 
  [SBJ\textsubscript{1} [V\textsubscript{2} \textit{at}-PP\textsubscript{3} ]\textsubscript{VP}]\textsubscript{4} & OBJ-of-PP\textsubscript{3} \textit{was not the target of the} V\textsubscript{2}-\textit{ing} \textit{motion}.  \\ \midrule
  \hline
  \textbf{Intransitive-Motion} &
[SBJ\textsubscript{1} [V\textsubscript{2} PP\textsubscript{3} ]\textsubscript{VP}]\textsubscript{4} & SBJ\textsubscript{1} V\textsubscript{2} \textit{in a static location}. \\ \hline
  \textbf{Let-Alone} &
XP\textsubscript{1} CONJ\textsubscript{2-3} XP\textsubscript{4-5} & \textit{If} XP\textsubscript{1} \textit{then not} XP\textsubscript{4-5} \\
  \hline
  \textbf{Resultative} &
  [SBJ\textsubscript{1} [V\textsubscript{2} OBJ\textsubscript{3} AP\textsubscript{4} ]\textsubscript{VP}]\textsubscript{5}& \textit{The} V\textsubscript{2}-\textit{ing} \textit{did not cause} OBJ\textsubscript{3} \textit{to become} AP\textsubscript{4}. \\ \midrule
  \hline
  \textbf{Way-Manner} &
[SBJ\textsubscript{1}\enspace[V\textsubscript{2}\enspace[PRON\textsubscript{3=1}]\enspace\textit{way}\textsubscript{4}\enspace]\textsubscript{OBJ5} PP\textsubscript{6}]\textsubscript{VP}]\textsubscript{7}

& SBJ\textsubscript{1} \textit{traveled} PP\textsubscript{6} \textit{without} V\textsubscript{2}-\textit{ing}. \\  &\\
  \hline
  
\bottomrule
\end{tabular}%
}
\caption{Example templates for Cxns and templatic constructional NLI hypotheses. All examples provided here have the contradiction relation. In these templates OBJ stands for a bare object, and OBL stands for an oblique, which is a prepositional phrase that introduces a recipient, goal, or result of the verb. AP, PP, and VP stand for adjective phrase, prepositional phrase, and verb phrase respectively.}
\label{tab:templates}
\end{table*}

\begin{table*}[ht]
\centering
\resizebox{\textwidth}{!}{%
\begin{tabular}{p{3cm}|p{2cm}|p{12.5cm}}
\toprule
\textbf{Cxn Name} & & \textbf{CxNLI (Exp. 1)} \\ \midrule
\hline
\textbf{Causative-With} & Premise &
  \textit{Freshly ground coffee beans filled the room with a seductive, earthy aroma.} \\ \midrule
  & Hypothesis & \textit{The room did not contain a seductive, earthy aroma.}\\
  & Relation & Contradiction \\
  \hline
\textbf{Caused-Motion} & Premise &
  \textit{I threw the stone across the river.} \\ \midrule
  & Hypothesis & \textit{I caused the stone to move across the river by throwing it.}\\
  & Relation & Entailment \\
  \hline
\textbf{Comparative-Correlative} & Premise &
  \textit{The more they work, the more I will pay them.} \\ \midrule
  & Hypothesis & \textit{Increasing the amount they work will increase the amount I pay them.}\\
  & Relation & Entailment \\
  \hline
\textbf{Conative} & Premise &
  \textit{I sipped at the Heineken.} \\ \midrule
  & Hypothesis & \textit{The Heineken was not the target of my sipping.}\\
  & Relation & Contradiction \\
  \hline
\textbf{Intransitive-Motion} & Premise &
  \textit{I ran around the track.} \\ \midrule
  & Hypothesis & \textit{I ran, staying in one place.}\\
  & Relation & Contradiction \\
  \hline
\textbf{Let-Alone} & Premise &
  \textit{It's unsurprising that such an attitude failed to produce competent screenwriters, let alone exciting ones.} \\ \midrule
  & Hypothesis & \textit{An attitude that produces exciting screenwriters can also produce competent ones.}\\
  & Relation & Entailment \\
  \hline
\textbf{Resultative} & Premise &
  \textit{The jackhammer pounded us deaf.} \\ \midrule
  & Hypothesis & \textit{We were completely deaf before the jackhammer pounded.}\\
  & Relation & Contradiction \\
  \hline
\textbf{Way-Manner} & Premise &
  \textit{I yawned my way back to the Narrow Neck.} \\ \midrule
  & Hypothesis & \textit{I traveled back to Narrow Neck without yawning.}\\
  & Relation & Contradiction \\
  \hline
\bottomrule
\end{tabular}%
}
\caption{Examples of templatically generated hypotheses for and resulting NLI tuples for Exp. 1 CxNLI.}
\label{tab:nli-examples_exp1}
\end{table*}

\begin{table*}[t]
\centering
\resizebox{\textwidth}{!}{%
\begin{tabular}{l|ll}
\textbf{CxN Name}                                                                   &            & \textbf{Example NLI Triple}                                                                                                                                                                       \\ \hline
\multirow{3}{*}{Intransitive+At}                                                           & Premise    & I watch the women, their legs crossed at the ankles, try to look as if they don't sweat at all.                                                                                                                                        \\
                                                                                    & Hypothesis & Their legs made a crossing motion towards their ankles.                                                                                                                                 \\
                                                                                    & Relation   & Contradiction                                                                                                                                                                                     \\ \hline
\multirow{3}{*}{\begin{tabular}[c]{@{}l@{}}Transitive+with\end{tabular}}          & Hypothesis & He hit the lamp with his head.                                                                                                                                    \\
                                                                                    & Premise    & He caused the lamp to contain his head.                                                                                                                          \\
                                                                                    & Relation   & Contradiction                                                                                                                                                                                         \\ \hline
\multirow{3}{*}{\begin{tabular}[c]{@{}l@{}}Intransitive+To\end{tabular}}     & Hypothesis & He spoke to the workers on the street corner.              \\
                                                                                    & Premise    & He changed locations by speaking. \\
                                                                                    & Relation   & Contradiction                                                                                                                                                                                       \\ \hline
\multirow{3}{*}{\begin{tabular}[c]{@{}l@{}}Ditransitive+NP,PP\end{tabular}}           & Hypothesis & I introduced her to my boss.                                                                                        \\
                                                                                    & Premise    & She remained in the same place.                                                                                                                                                                  \\
                                                                                    & Relation   & Neutral                                                                                                                                                                                   \\ \hline
\multirow{3}{*}{Depictive}                                                        & Hypothesis & A famous emperor buried China’s scholars alive with their books.                                                                                                                                   \\
                                                                                    & Premise    & Burying caused the scholars to become alive.                                                                                                                                        \\
                                                                                    & Relation   & Contradiction                                                                                                                                                                                       
\end{tabular}%
}
\caption{Example Exp. 2 NLI triples corresponding to each of the 5 constructions evaluated in the Exp. 2 Cxn NLI dataset. 5 of the 8 Exp. 1 Cxns are included in this dataset; limited to those Exp. 1 constructions with syntactically identical counterpart Cxns that have different meanings. }
\label{tab:cnli-9-cxn-adv-examples}
\end{table*}


\section{Prompt Variation Experiments}
\label{sec:appendix_prompt_variation}
For each experiment, we choose one setting to observe the impact of prompt variations on performance.\footnote{For NLI we choose the 3-shot CxNLI setting.} These prompt variations were explored in two sequential stages; changing the content of the prompt and changing the phrasing of the prompt. Changing the content of the prompt involved giving more details about the task, giving fewer details about the task and experimenting with the specific instructions. After selecting the best-performing prompt content, the prompt was rephrased using an LLM and variations of the prompt with new wording were used to check the robustness of the model output by observing how small variations in the prompts affect performance. We report an example of each of our prompts for variations in content in Table \ref{tab:prompts12}. Results for our prompt variation experiments are shared in Tables \ref{tab:exp1scores_prompt} and \ref{tab:exp2bscores_prompt}.


\begin{table*}
    \small
    \centering
    \begin{tabular}{c|p{13cm}}
    \textbf{Prompt Type} & \textbf{Prompt} \\
        \hline
        Variation 1 & You are the world's best annotator. Your task is to read sentences from a dataset, presented as the Premise in a set of triples for the Natural Language Inference (NLI) task. Also known as Recognizing Textual Entailment (RTE), NLI involves determining the inference relation between two short, ordered texts: entailment, contradiction, or neutral. Next, you will identify the Relation between the Premise and the Hypothesis, which indicates the type of entailment between the two sentences. We use numerical coding, also listed in your annotation spreadsheet as a reminder:\\
        &0 – entailment – The hypothesis must be true given the premise\\
        &1 – neutral – The hypothesis may or may not be true given the premise\\
        &2 – contradiction – The hypothesis must not be true given the premise\\
        &Output a single numerical value between 0, 1, or 2, corresponding to the associated relation. Output a single number only and nothing else.\\
        
        \hline

        Variation 2 & You are the world's best annotator. You are tasked with annotating a triple for Natural Language Inference. You must determine the inference relation between the Premise and the Hypothesis by selecting one of three numerical codes that reflect the relationship:\\
        &0 – Entailment: The Hypothesis is definitely true given the Premise.\\
        &1 – Neutral: The Hypothesis may or may not be true given the Premise.\\
        &2 – Contradiction: The Hypothesis cannot be true given the Premise.\\
        &Output a single numerical value between 0 and 2 inclusive, corresponding to the associated relation.\\

        \hline
        
        Variation 3 & You are the best at understanding language inference based on Cxn grammar. You are tasked with annotating a triple for Natural Language Inference. You must determine the inference relation between the premise and the hypothesis by selecting one of three numerical codes that reflect the relationship:\\
        &0 – entailment – The hypothesis must be true given the premise\\
        &1 – neutral – The hypothesis may or may not be true given the premise\\
        &2 – contradiction – The hypothesis must not be true given the premise\\
        &Output a single numerical value between 0, 1, or 2, corresponding to the associated relation. Output a single number only and nothing else.\\

        \hline

        Variation 4 & You are the world's best annotator. Your task is to read sentences from a dataset, provided as the Premise in a set of triples for the Natural Language Inference (NLI) task. Also called Recognizing Textual Entailment (RTE), NLI requires determining the inference relation between two short, ordered texts: entailment, contradiction, or neutral. Your next step is to identify the Relation between the Premise and the Hypothesis, specifying the type of entailment between the two sentences. We use the following numerical coding:\\
        &0 – entailment – The hypothesis must be true given the premise\\
        &1 – neutral – The hypothesis may or may not be true given the premise\\
        &2 – contradiction – The hypothesis must not be true given the premise\\
        &Output a single numerical value between 0 and 2 inclusive, corresponding to the associated relation.\\

    \end{tabular}
    \caption{Prompt variations for Exp. 1 - CxNLI and  Exp. 2 - CxNLI-Distinction}
    \label{tab:prompts12}
\end{table*}   

\begin{table}[!h]
\centering
\begin{tabular}{ccc}
\hline
\textbf{Prompt Type}       & \multicolumn{2}{c}{\textbf{Accuracy}}\\ 
 & GPT-3.5 & GPT-4o\\
\hline

Variation 1 & 0.74 & 0.95\\
Variation 2 & 0.79 & 0.92\\
Variation 3 & 0.76 & 0.94\\
Variation 4 & 0.74 & 0.93\\
\hdashline
Best Variation Rephrase 1 & 0.79 & 0.95\\
Best Variation Rephrase 2 & 0.67 & 0.95\\
Best Variation Rephrase 3 & 0.69 & 0.96\\

\end{tabular}%
\caption{Prompt Variation Results for Exp. 1 - CxNLI in the three-shot setting.}
\label{tab:exp1scores_prompt}
\end{table}

\begin{table}[!h]
\centering
\begin{tabular}{ccc}
\hline
\textbf{Prompt Type}       & \multicolumn{2}{c}{\textbf{Accuracy}}\\ 
 & GPT-3.5 & GPT-4o\\
\hline
Variation 1 & 0.24 & 0.53\\
Variation 2 & 0.28 & 0.46\\
Variation 3 & 0.31 & 0.57\\
Variation 4 & 0.26 & 0.54\\
Best Variation Rephrase 1 & 0.31 & 0.57\\
Best Variation Rephrase 2 & 0.26 & 0.55\\
Best Variation Rephrase 3 & 0.23 & 0.54\\
\end{tabular}%
\caption{Prompt Variation Results for Exp. 2 CxNLI-Distinction in the three-shot setting.}
\label{tab:exp2bscores_prompt}
\end{table}

\section{Chain-of-Thought Results for all Experiments}
\label{sec:appendix_cot}
We replicate each of our main experiments with the addition of Chain-of-Thought (CoT) to observe changes in performance and inspect any errors in the model's reasoning steps. For CoT thought prompting we do not provide examples of reasoning in the prompt, instead we simply ask the model to explain "step by step". The results for each experiment have been shared in tables \ref{tab:exp1cotscores} and \ref{tab:exp2bcotscores}.

\begin{table}[!h]
\centering
\begin{tabular}{p{1.7cm}p{1.2cm}p{1.4cm}p{1.3cm}}
\hline
\textbf{Setting} & \textbf{IC Data} & \multicolumn{2}{c}{\textbf{Accuracy}}\\ 
 & & GPT-3.5 & GPT-4o\\
\hline
Zero-shot & None & 0.60 & 0.89 \\
One-shot & CxNLI & 0.66 & 0.89 \\
Three-shot & CxNLI & 0.71 & 0.92 \\
One-shot & SNLI & 0.63 & 0.89\\
Three-shot & SNLI & 0.66 & 0.91 \\
\end{tabular}%
\caption{Results for Exp. 1 - Cxn NLI with Chain-of-Thought, "IC Data" refers to the type of data used as in-context examples.}
\label{tab:exp1cotscores}
\end{table}


\begin{table}[!h]
\centering
\begin{tabular}{ccccc}
\hline
\textbf{Setting} &  \textbf{IC Data}     & \multicolumn{2}{c}{\textbf{Accuracy}} \\ 
 & & GPT-3.5 & GPT-4o\\
\hline
Zero-shot & None & 0.29 & 0.43\\
One-shot & CxNLI & 0.29 &  0.43\\
Three-shot & CxNLI & 0.35 & 0.46\\
One-shot & SNLI & 0.28 & 0.43\\
Three-shot & SNLI & 0.31 & 0.45\\
\end{tabular}%
\caption{Results for Exp. 2 with the CxNLI-Distinction data as the test and Chain-of-Thought, "IC Data" refers to the type of data used as in-context examples.}
\label{tab:exp2bcotscores}
\end{table}

\section{Model Parameters and Hyperparameters}

We use the default hyperparameters for all models including GPT-4o (unreported parameter size), GPT-3.5 (175B parameters), GPT-o1 (unreported parameter size), LLaMA 3 8B (8B parameters), LLaMA 3 70B (70B parameters) though to ensure maximal reproducibility of our work, we set the temperature of model responses to 0 apart from GPT-o1 which only acceptd a temperature of 1. Our GPT experiments were done through the OpenAI API. The total cost of our experiments was approximately \$250 USD. Our LLaMA experiments were run through the replicate API and the experiments cost approximately \$25 USD. 

\section{Constructional Natural Language Inference Annotation Guidelines}
\label{appendix:nli-guidelines}

\textit{Here we provide the exact instructions given to people to annotate the NLI datasets.}  

\noindent We have developed a dataset of sentences featuring different linguistic ``constructions''---pairings of form and meaning. The constructions exemplified in this dataset range from purely substantive (the words filling the constructional slots are fixed), such as the \textit{Much-less} construction, e.g., ``He won’t eat shrimp, much less squid;'' to purely schematic (the words filling the constructional slots can vary, but fulfill some general semantic and syntactic requirements), such as the \textit{Caused-Motion} construction, e.g., ``She blinked the snow off her eyelashes.'' It’s okay if you aren’t familiar with this terminology or the idea of these constructions!

\subsection*{Task Overview}

Your job is to read the sentences from this dataset, which are presented as the \textbf{Premise} in a set of triples for the Natural Language Inference (NLI) task. Also known as Recognizing Textual Entailment (RTE), NLI is the task of determining the inference relation between two (short, ordered) texts: entailment, contradiction, or neutral \cite{maccartney-manning-2008-modeling}.

\begin{itemize}
  \item \textbf{Premise:} A man inspects the uniform of a figure in some East Asian country.
  \item \textbf{Hypothesis:} The man is sleeping.
\end{itemize}

Then, you will fill in the \textbf{Relation} between the Premise and the Hypothesis, which indicates the kind of entailment between the two sentences. We are using numerical coding, listed below and in your annotation spreadsheet:

\begin{itemize}
  \item[0 --] \textbf{entailment} – The hypothesis must be true given the premise.
  \item[1 --] \textbf{neutral} – The hypothesis may or may not be true given the premise.
  \item[2 --] \textbf{contradiction} – The hypothesis must not be true given the premise.
\end{itemize}

So for the example above, the correct answer would be:

\begin{quote}
\textbf{2 – contradiction} – If the man is inspecting a uniform, then it must not be true that the man is sleeping.
\end{quote}

The two sentences describe the \textit{same scenario}. Entities mentioned in both premise and hypothesis refer to the same thing; e.g., ``the man'' refers to the same individual. The hypothesis does not describe a different time.

If you encounter unfamiliar words, you may consult a dictionary. However, it is not expected or encouraged that you would have to “do research” into a topic in order to determine a relation between a premise and hypothesis. Instead, you should rely on common sense and your understanding of the words. 

You will be completing these annotations in a spreadsheet like what is shown below, where there is a relation space available below a given premise/hypothesis pair. In the cell to the right of “relation”, you will provide the appropriate relation number.  This space should include nothing except for the numbers 0, 1, or 2. 

If you would like to note instances that are problematic, please add a Notes column to the right of Annotation Target and make your note as relevant to the right of the premise, hypothesis or relation. 

There is also a space in column L where you can note the start and end times of each annotation session, or “sitting.” Please kindly track about how many judgments you are able to do in each sitting, so we can get a sense of how long the annotation task takes. 

There are about 100 distinct NLI judgments or triples per annotation spreadsheet. Once you have completed all the relation annotations, please save and send the spreadsheet back to me: Claire.n.bonial.civ@army.mil. 

Get your inference hat on. Happy annotating!

\end{document}